\documentclass{article}

\usepackage{PRIMEarxiv}

\usepackage[utf8]{inputenc}
\usepackage[T1]{fontenc}
\usepackage[hidelinks]{hyperref}
\usepackage{url}
\usepackage{booktabs}
\usepackage{amsfonts,amsmath,amssymb}
\usepackage{nicefrac}
\usepackage[nopatch=footnote]{microtype}
\usepackage{fancyhdr}
\usepackage{setspace}
\usepackage{graphicx}
\usepackage{float}
\graphicspath{{media/}}

\newcommand{\orcidicon}[1]{%
  \href{https://orcid.org/#1}{%
    \includegraphics[height=1.5ex]{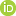}%
  }%
}

\usepackage{fancyhdr}
\title{Indian Sign Language Detection for Real-Time Translation using Machine Learning
\thanks{\textit{Citation}: Rajat Singhal, Jatin Gupta, Akhil Sharma, Anushka Gupta, Navya Sharma. "Indian Sign Language Detection for Real-Time Translation using Machine Learning". Proceedings of the 6\textsuperscript{th} International Conference on Recent Advances in Information Technology (RAIT), 2025, IEEE. DOI: \href{https://doi.org/10.1109/RAIT65068.2025.11089142}{10.1109/RAIT65068.2025.11089142}}
}

\author{
Rajat Singhal \orcidicon{0009-0002-8017-2306} \\
Department of Computer Science and Engineering \\
Sharda University, Greater Noida, India \\
\texttt{singhalrajat354@gmail.com} \\
\And
Jatin Gupta \orcidicon{0009-0002-9504-7487} \\
Department of Computer Science and Engineering \\
Sharda University, Greater Noida, India \\
\texttt{jatingupta261001@gmail.com} \\
\And
Akhil Sharma \orcidicon{0009-0001-1490-4022} \\
Department of Computer Science and Engineering \\
Sharda University, Greater Noida, India \\
\texttt{sharmaakhil944@gmail.com} \\
\And
Anushka Gupta \\
Department of Computer Science and Engineering \\
Babu Banarasi Das University, Lucknow, India \\
\texttt{anushkag472004@ieee.org} \\
\And
Navya Sharma\thanks{Corresponding author: Navya Sharma (email: \texttt{navyasav06@gmail.com}).}  \hspace{0.075cm} \orcidicon{0009-0008-5977-4146} \\
Department of Computer Science and Engineering \\
Sharda University, Greater Noida, India \\
\texttt{navyasav06@gmail.com}
}

\begin{document}
\maketitle

\begin{abstract}
Gestural language is used by deaf and mute communities to communicate through hand gestures and body movements that rely on visual and spatial patterns known as sign languages. Sign languages, which rely on visual-spatial patterns of hand gestures and body movements, are the primary mode of communication for deaf and mute communities worldwide. Effective communication is fundamental to human interaction, yet individuals in these communities often face significant barriers due to a scarcity of skilled interpreters and accessible translation technologies. This research specifically addresses these challenges within the Indian context by focusing on Indian Sign Language (ISL). By leveraging machine learning, this study aims to bridge the critical communication gap for the deaf and hard-of-hearing population in India, where technological solutions for ISL are less developed compared to other global sign languages. We propose a robust, real-time ISL detection and translation system built upon a Convolutional Neural Network (CNN). Our model is trained on a comprehensive ISL dataset and demonstrates exceptional performance, achieving a classification accuracy of 99.95\%. This high precision underscores the model's capability to discern the nuanced visual features of different signs. The system's effectiveness is rigorously evaluated using key performance metrics, including accuracy, F1 score, precision, and recall, ensuring its reliability for real-world applications. For real-time implementation, the framework integrates MediaPipe for precise hand tracking and motion detection, enabling seamless translation of dynamic gestures. This paper provides a detailed account of the model's architecture, the data preprocessing pipeline, and the classification methodology. The research elaborates the Model architecture, preprocessing, and classification methodologies for enhancing communication in deaf and mute communities.
\end{abstract}

\keywords{Indian Sign Language \and Machine Learning \and Real-Time \and Translation \and Deep Learning \and Deaf and Mute \and Convolutional Neural Network (CNN)}

\twocolumn

\section{Introduction}
Sign language serves as a vital mode of communication for individuals who are deaf or speech-impaired, enabling them to convey thoughts, emotions, and needs through gestures. It is more than a tool; it’s a cultural and linguistic bridge connecting them with the hearing world. Different regions have distinct sign languages tailored to their unique cultural and linguistic nuances. For instance, Indian Sign Language (ISL) is integral to India’s deaf and mute communities, helping millions overcome communication barriers \cite{Mishra2023}.

Despite its significance, sign language faces challenges in mainstream adoption. A major issue is the general public’s limited awareness and understanding of it. Unlike spoken languages, sign languages like ISL remain underrepresented in educational systems, technological advancements, and daily interactions~\cite{Rokade2017}. This lack of awareness, coupled with a shortage of trained interpreters, often isolates the deaf and mute community in India \cite{Malla2024}.

Emerging technologies, particularly artificial intelligence (AI) and machine learning (ML), hold promise for bridging this gap. These technologies have been leveraged to develop sign language translation systems, using computer vision and gesture recognition to convert signs into readable or audible formats. However, most innovations focus on widely recognized systems like American Sign Language (ASL), leaving languages like ISL underexplored. The unique syntax and structure of ISL require specialized research to enhance its translation capabilities, ensuring inclusivity for India’s deaf and hard-of-hearing population \cite{Aggarwal2023}.

Further integration with IoT (Internet of Things) technologies can revolutionize real-time sign language translation. IoT-enabled systems can facilitate seamless communication by integrating devices and networks, making translation tools more accessible and efficient. By adopting these advancements, we can work toward greater inclusion and empowerment for sign language users worldwide.
\begin{figure}[h!]
    \centering
    \includegraphics[width=1\linewidth]{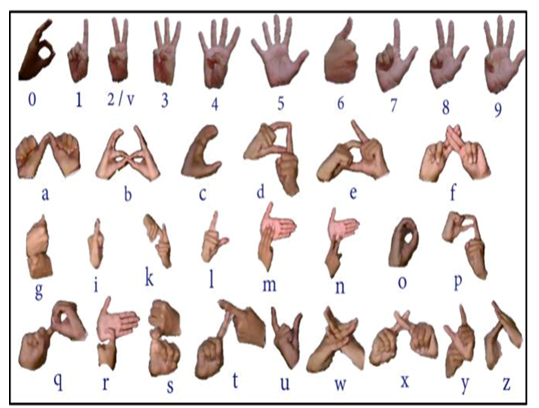}
    \caption{Sample Image from Dataset}
    \label{fig:dataset}
\end{figure}

This paper presents the development of a real-time recognition system for 35 distinct hand gestures of Indian Sign Language (ISL). These gestures include representations of alphabets and commonly used words in daily communication. A detailed review of existing research on Indian Sign Language is outlined in the next section, highlighting relevant studies and advancements.

The proposed methodology is elaborated in Section III, focusing on dataset collection and preprocessing techniques. This includes capturing ISL gestures and refining them for optimal use in model training. Section IV delves into the recognition methods employed to interpret ISL gestures accurately, describing the models and techniques utilized.

Finally, Section V provides an analysis of the experimental results obtained from implementing these methods. This section evaluates the system’s performance and discusses the outcomes, offering insights into the effectiveness of the proposed techniques.

\section{Related Work}
Indian Sign Language (ISL) recognition is a critical field of research aimed at addressing communication challenges faced by the deaf and mute communities. By leveraging machine learning and deep learning techniques, researchers have sought to enhance the accuracy and efficiency of ISL recognition systems.

Deep learning models have shown significant promise in ISL analysis. For example, studies comparing various architectures, such as pre-trained VGG16, VGG16 with adaptive learning, and hierarchical neural networks, demonstrate that hierarchical models outperform others, achieving an impressive accuracy of 98.52\% for single-hand gestures and 97\% for both hands combined~\cite{Sharma2021}. Another investigation into convolutional neural networks (CNNs) revealed that out of 50 CNN configurations tested, the highest training accuracies were 99.72\% for color images and 99.90\% for grayscale images \cite{Wadhawan2020}.

Feature extraction techniques have also been widely explored in ISL research. Methods such as Oriented FAST and Rotated BRIEF (ORB) and Scale Invariant Feature Transform (SIFT) have been analyzed for their performance in conjunction with classifiers like Support Vector Machines (SVM). Additional approaches, including edge detection, object removal, and word packing, have been employed to improve recognition capabilities \cite{Munnaluri2022}.

Hybrid neural network architectures have shown exceptional accuracy for both static and dynamic gestures. For instance, combining 3D Convolutional Nets with atrous convolution for static gestures, and employing a hybrid model integrating semantic multi-mark feature detection with physical sequential feature extraction for dynamic gestures, achieved accuracy rates of 99.76\% and 99.85\%, respectively \cite{Rajalakshmi2023}.

In the realm of traditional machine learning, algorithms like Random Forest have demonstrated strong performance, with studies reporting an accuracy of 98.44\% when evaluated against five other algorithms \cite{Potnis2021}. Similarly, MATLAB-based experiments have reported accuracy rates ranging from 92\% to 100\% for single and dual-hand gesture recognition \cite{Dutta2017}.

Real-time ISL recognition systems have also been developed to support applications in gesture-controlled robotics, automated systems, and human-computer interactions. Techniques like fuzzy c-means clustering have been pivotal in these advancements \cite{Mariappan2019}. However, challenges such as variability in lighting, backgrounds, and signer differences continue to pose significant obstacles. Researchers have addressed these issues by simulating such variations during model testing to improve robustness \cite{Sharma2021, Rajalakshmi2023}.

Furthermore, the integration of ISL recognition systems into publicly accessible platforms and educational tools has the potential to greatly enhance the lives of individuals in the deaf and mute communities. These platforms aim to provide not only gesture recognition but also resources for education and skill development, thereby improving employment prospects \cite{Mistry2021}.

This study aims to contribute to the field of ISL recognition by introducing a deep convolutional neural network (CNN) architecture optimized for accurate classification of ISL gestures. Unlike prior work that predominantly focuses on languages like American Sign Language (ASL), this research is tailored for ISL, addressing the critical need for robust communication solutions in India. The proposed model employs a multi-step image preprocessing pipeline and a CNN-based system to effectively capture and classify visual patterns unique to ISL gestures.

\section{Proposed work}

\subsection{Dataset collection }\label{AA}
The dataset plays a pivotal role in developing machine learning models, particularly for gesture recognition, as it significantly influences the system's performance and accuracy. For this research, the Indian Sign Language (ISL) Dataset has been utilized. It comprises 35 distinct hand gestures, including digits from 0 to 9 and alphabets from A to Z (as shown in Figure \ref{fig:dataset}). Each class includes 1,000 images captured from various perspectives, resulting in a total of 42,743 files distributed across 35 classes.

\subsection{Preprocessing}
\begin{figure}[h!]
\centering
\includegraphics[width=0.4\linewidth]{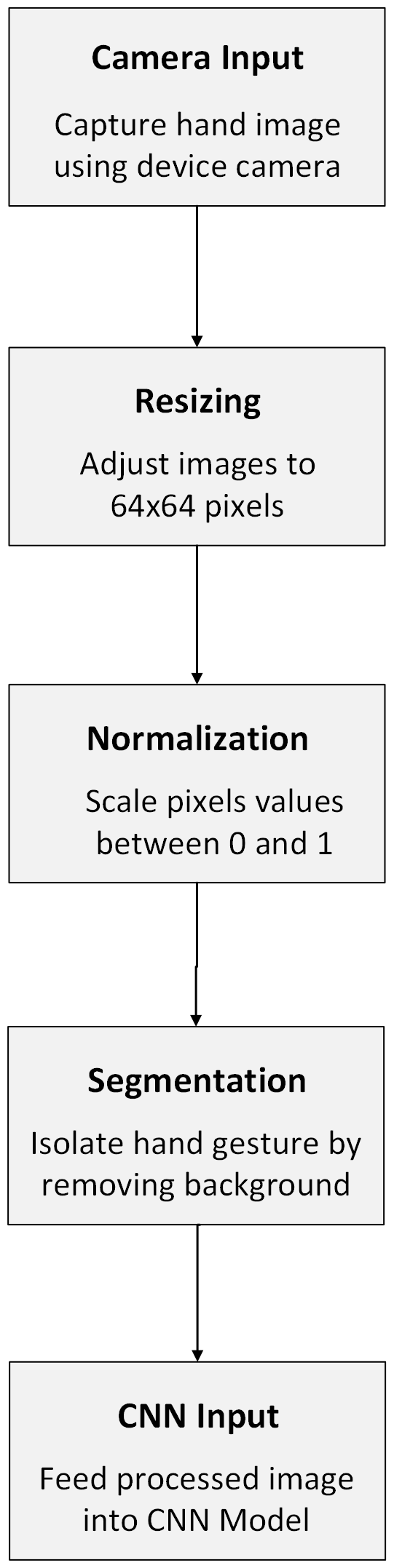}
\caption{Preprocessing Workflow}
\label{fig:preprocessing}
\end{figure}

Data preprocessing, as depicted in Figure \ref{fig:preprocessing}, is an essential step to ensure that machine learning models can learn effectively from input data. Its primary goal is to clean, structure, and organize the data while preserving the unique features of Indian Sign Language (ISL) gestures. This project employed several preprocessing techniques to improve the quality of the training data and enhance model accuracy.

\subsubsection{\textbf{Greyscale Conversion}}
The first preprocessing step involved converting RGB images to grayscale. This step is vital in image analysis as it reduces data dimensionality, thereby speeding up processing and simplifying subsequent analysis. The grayscale conversion follows the luminance model, expressed in Equation~\ref{eq:grey}:

\begin{equation}
\label{eq:grey}
Y = 0.299 \cdot R + 0.587 \cdot G + 0.114 \cdot B
\end{equation}

\noindent Here, $Y$ is the grayscale intensity, and $R$, $G$, and $B$ denote the intensities of the red, green, and blue channels, respectively.

\subsubsection{\textbf{Binary Thresholding}}
Binary thresholding involves transforming a grayscale image into a binary one by applying a set threshold value. This process emphasizes contrast by separating the object of interest from the background. A threshold value of 90 was chosen to highlight specific features effectively. The binary conversion is mathematically represented in Equation~\ref{eq:binary}.

\begin{equation}
\label{eq:binary}
f(x, y) =
\begin{cases}
0, & \text{if } f(x, y) < T \\
255, & \text{otherwise}
\end{cases}
\end{equation}

\noindent \textbf{where:}
\begin{itemize}
\item  f(x, y)  is the pixel intensity at the given coordinates,
\item  T  represents the threshold value.
\end{itemize}

\subsubsection{\textbf{Canny Edge Detection}}
\begin{figure}[h!]
\centering
\includegraphics[width=0.5\linewidth]{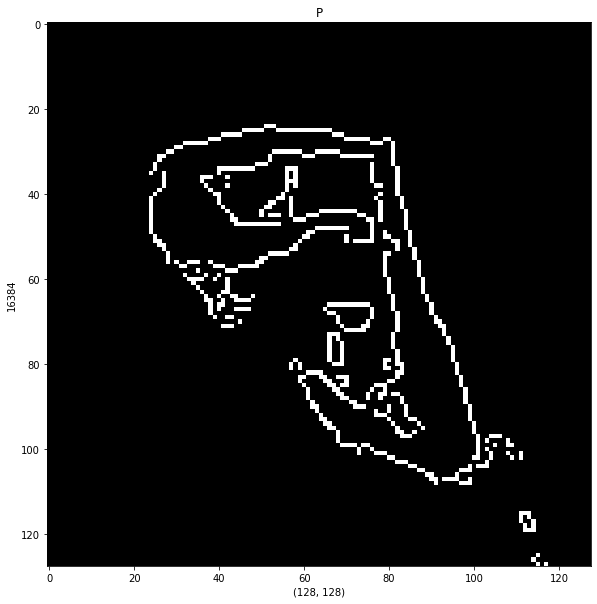}
\caption{Canny Edge Detection}
\label{fig:canny-edge-detection}
\end{figure}

Edge detection was performed using the Canny edge detector, as illustrated in Figure \ref{fig:canny-edge-detection}. This technique computes the gradient of each pixel to determine edge strength and orientation, enhancing the contrast between edges and the background. By leveraging intensity gradients, it improves the clarity of edges, which is crucial for effective model learning. The edge gradient and orientation of each pixel are calculated as shown in Equation~\ref{eq:edge} and Equation~\ref{eq:pix}, respectively.

\begin{equation}
\label{eq:edge}
\text{EdgeGradient} (G) = \sqrt{G_x^2 + G_y^2}
\end{equation}

\begin{equation}
\label{eq:pix}
\text{Angle} (\theta) = \tan^{-1} \left( \frac{G_y}{G_x} \right)
\end{equation}

The Sobel kernel was applied horizontally and vertically to calculate the first derivatives (\(G_x\)) and (\(G_y\)), which represent the gradient in the respective directions.

\subsection{Dataset Splitting}
To facilitate effective training and evaluation of the machine learning model, the dataset was divided into two distinct subsets: a training set and a testing set. This partitioning enabled the model to optimize its learning process using the training data while being evaluated on unseen testing data. Such a strategy helps minimize overfitting and provides a reliable assessment of the model’s applicability in real-world scenarios.

\subsubsection{\textbf{Training Set}}
The training set serves as the primary dataset used for teaching the model. It enables the model to recognize patterns, features, and relationships within the data, forming the foundation of its learning process. The quality and comprehensiveness of the training set directly affect the model’s ability to generalize to unseen data.

\subsubsection{\textbf{Testing Set}}
The testing set is entirely separate from the training data and is reserved exclusively for performance evaluation. By using unseen data for testing, the model’s performance can be assessed impartially, offering an unbiased prediction of its effectiveness in real-world applications.

Dividing the dataset into training and testing sets is crucial for ensuring robustness and avoiding overfitting, where the model memorizes the training data rather than learning meaningful patterns. A common approach is to split the data in an 80:20 ratio, with 80\% used for training and 20\% for testing. However, the exact proportions may vary depending on the dataset size and the project’s specific requirements.

\section{Classification}
The proposed system for Indian Sign Language (ISL) recognition leverages Convolutional Neural Networks (CNNs) due to their demonstrated effectiveness in image classification tasks and their unique ability to extract meaningful features from visual data.

\subsection{Convolutional Neural Network (CNN)}\label{AA}
Convolutional Neural Networks (CNNs) are a specialized class of deep learning models designed for image and pattern recognition. They excel at identifying spatial hierarchies in images, capturing both global patterns (e.g., shapes and structures) and local features (e.g., edges, textures, and angles). This makes CNNs particularly well-suited for Indian Sign Language (ISL) recognition, where precise analysis of hand gestures and movements is essential \cite{Jha2022}.

The features which contribute to the success of CNNs include:
\begin{itemize}
	\item Convolutional Layers: These apply filters (kernels) to the input image, enabling the model to detect significant features and create a hierarchical representation of the data.
	\item Pooling Layers: Techniques like max pooling reduce the spatial dimensions of feature maps while retaining the most critical information. For instance, max pooling selects the maximum value within a specific region, helping to highlight essential patterns and reduce computational complexity.
	\item Fully Connected Layers: After feature extraction, these layers map the learned features to the target labels, forming predictions by modeling the relationship between features and output classes \cite{Jha2023}.
\end{itemize}

The network employs nonlinear activation functions like ReLU (Rectified Linear Unit), which allow it to model complex relationships effectively. This adaptability makes CNNs an ideal choice for ISL recognition, as they robustly learn hierarchical gesture features and are resilient to variations in lighting, rotation, and scaling \cite{Adithya2020}.

\subsection{Training the Model}\label{AA}
Training a Convolutional Neural Network (CNN) requires feeding the input data, such as images of gestures, into the model. This data is used to adjust the model’s parameters through an iterative process, where the goal is to minimize the error between predicted and actual outputs. Optimization techniques like stochastic gradient descent (SGD) are commonly applied to achieve this, while methods such as early stopping and validation at regular intervals help prevent overfitting and ensure the model performs at its best.

\subsection{CNN Architecture}\label{AA}
The architecture of the CNN model is designed using a combination of convolutional layers, pooling layers, dropout layers, and fully connected (dense) layers. Each layer plays a unique role in enhancing the model’s performance and its ability to generalize. The first convolutional layer, equipped with 24 filters (3x3 kernel size) and ReLU activation, extracts basic features like edges and simple textures from the input images. Batch normalization is applied in this layer to improve training stability. Following this, a max pooling layer is used to reduce the spatial dimensions by 50\%, ensuring that the most significant features are preserved.

To capture increasingly complex features, three additional convolutional layers are added, with 64, 128, and 256 filters respectively. Each of these layers also uses ReLU activation to allow the network to learn complex hierarchical patterns in the gesture images. After each convolutional block, max pooling further helps in down-sampling the feature maps to retain only the most critical information. The model then passes the output through a fully connected layer consisting of 2,352 units, with ReLU activation applied to smooth the results. Finally, the output is passed through a softmax activation function in the last dense layer, which calculates the probability distribution for each gesture class based on the model’s learned representations.
\begin{table}[h]
\centering
\scriptsize
\caption{Model Architecture Summary}
\begin{tabular}{|l|l|l|}
\hline
\textbf{Layer (Type)} & \textbf{Output Shape} & \textbf{Param \#} \\ \hline
conv2d\_6 & (None, 254, 254, 24) & 240 \\ \hline
batch\_normalization\_1  & (None, 254, 254, 24) & 96 \\ \hline
max\_pooling2d\_5  & (None, 127, 127, 24) & 0 \\ \hline
conv2d\_7  & (None, 127, 127, 64) & 13,888 \\ \hline
dropout\_5  & (None, 127, 127, 64) & 0 \\ \hline
max\_pooling2d\_6  & (None, 63, 63, 64) & 0 \\ \hline
conv2d\_8  & (None, 63, 63, 64) & 36,928 \\ \hline
dropout\_6  & (None, 63, 63, 64) & 0 \\ \hline
max\_pooling2d\_7  & (None, 31, 31, 64) & 0 \\ \hline
conv2d\_9  & (None, 31, 31, 128) & 73,856 \\ \hline
conv2d\_10  & (None, 31, 31, 128) & 147,584 \\ \hline
dropout\_7  & (None, 31, 31, 128) & 0 \\ \hline
max\_pooling2d\_8  & (None, 15, 15, 128) & 0 \\ \hline
conv2d\_11  & (None, 15, 15, 256) & 295,168 \\ \hline
dropout\_8  & (None, 15, 15, 256) & 0 \\ \hline
max\_pooling2d\_9  & (None, 7, 7, 256) & 0 \\ \hline
flatten\_1 & (None, 12544) & 0 \\ \hline
dense\_2  & (None, 2352) & 29,505,840 \\ \hline
dropout\_9  & (None, 2352) & 0 \\ \hline
dense\_3  & (None, 35) & 82,355 \\ \hline
\textbf{Total Parameters} & \multicolumn{2}{c|}{\textbf{30,155,955}} \\ \hline
\textbf{Trainable Parameters} & \multicolumn{2}{c|}{\textbf{30,155,907}} \\ \hline
\textbf{Non-trainable Parameters} & \multicolumn{2}{c|}{\textbf{48}} \\ \hline
\end{tabular}
\label{tab:model_architecture}
\end{table}

\subsection{Evaluation Metrics}

\subsubsection{\textbf{Recall}}
Recall measures the model’s ability to correctly identify all actual positive instances, emphasizing the reduction of false negatives. A higher recall value indicates fewer relevant instances being missed. The formulation is shown in Equation~\ref{eq:recall}.

\begin{equation}
\label{eq:recall}
\text{Recall} = \frac{TP}{TP + FN}
\end{equation}
where $TP$ denotes true positives, and $FN$ stands for false negatives.

\subsubsection{\textbf{Precision}}
Precision indicates the accuracy of positive predictions by calculating the proportion of correctly identified positive cases out of all cases predicted as positive. A higher precision indicates fewer false positives. Equation~\ref{eq:precision} details the computation used to evaluate recall performance.

\begin{equation}
\label{eq:precision}
\text{Precision} = \frac{TP}{TP + FP}
\end{equation}
where $TP$ represents true positives, and $FP$ refers to false positives.

\subsubsection{\textbf{Accuracy}}
Accuracy gives a broad view of the model’s overall performance by evaluating the proportion of correct predictions—both positive and negative—compared to the total predictions made (see Equation~\ref{eq:acc}). It’s useful for assessing general model effectiveness but can be misleading for imbalanced datasets.

\begin{equation}
\label{eq:acc}
\text{Accuracy} = \frac{TP + TN}{TP + TN + FP + FN}
\end{equation}
where $TP$ and $TN$ indicate true positives and true negatives, respectively, while $FP$ and $FN$ represent false positives and false negatives.

\subsubsection{\textbf{F1 Score}}
The F1 Score combines Precision and Recall into a single metric, providing a balance between them. See Equation~\ref{eq:f1} for the complete formulation. This score is especially helpful when dealing with imbalanced datasets, where both false positives and false negatives need to be minimized. 

\begin{equation}
\label{eq:f1}
\text{F1 Score} = 2 \times \frac{\text{Precision} \times \text{Recall}}{\text{Precision} + \text{Recall}}
\end{equation}

\section {Experiments and Results}
The proposed CNN model was evaluated using a dedicated test set to assess its capability in classifying Indian Sign Language (ISL) gestures. The evaluation revealed that the model achieved a remarkable accuracy of approximately 99.95\%, showcasing its proficiency in accurately identifying the complex spatial patterns associated with ISL gestures. This high accuracy demonstrates the robustness and effectiveness of the CNN in handling intricate visual tasks.

To further validate the performance, a comparative analysis was conducted against traditional models like VGG16, K-Nearest Neighbors (KNN), and Logistic Regression. The CNN demonstrated significantly superior results, affirming the advantages of deep learning methods over conventional classifiers for visual data like ISL gestures.

\begin{figure}[H]
    \centering
    \includegraphics[width=1\linewidth]{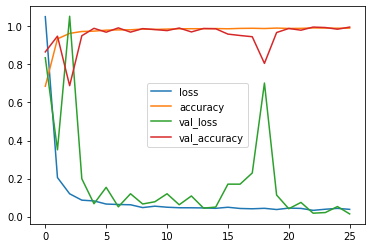}
    \caption{CNN Model History}
    \label{fig:CNN Model History}
\end{figure}

\noindent Figure \ref{fig:CNN Model History} illustrates the training history of the CNN model over 25 epochs, detailing metrics such as training loss, training accuracy, validation loss (\textit{val\_loss}), and validation accuracy (\textit{val\_accuracy}) .

\begin{figure}[H]
    \centering
    \includegraphics[width=1\linewidth]{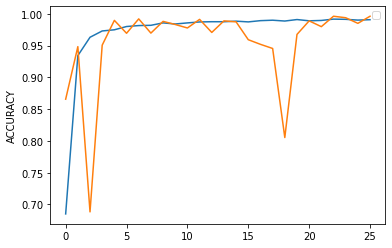}
    \caption{CNN Model Accuracy Curve}
    \label{fig:Model Accuracy}
\end{figure}
\noindent Figure \ref{fig:Model Accuracy} presents the model’s accuracy progression over 25 epochs, highlighting the achievement of 99.95\% accuracy in a few epochs.
\begin{figure}[H]
    \centering
    \includegraphics[width=1\linewidth]{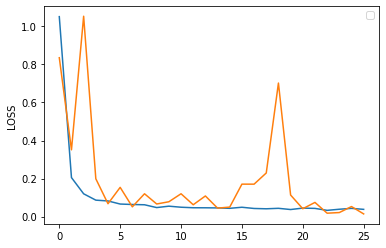}
    \caption{CNN Model Loss Curve}
    \label{fig:loss-graph}
\end{figure}
\noindent Figure \ref{fig:loss-graph} depicts the model’s loss curve over 25 epochs, showing consistent optimization and convergence.

\begin{table}[h!]
\caption{Model Architectures and Their Performance on the ISL Dataset}
\small
\centering
\renewcommand{\arraystretch}{1.5} 
\resizebox{\linewidth}{!}{ 
\begin{tabular}{|c|c|c|c|}
\hline
\textbf{S.No} & \textbf{Architecture} & \textbf{Dataset} & \textbf{Accuracy} \\ \hline
1 & VGG16~\cite{10649094} & ISL & 95.56\% \\ \hline
2 & AlexNet~\cite{10649094} & ISL & 93.9\% \\ \hline
3 & CNN & ISL & 99.95\% \\ \hline
\end{tabular}
}
\label{tab:model-performance}
\end{table}

Table \ref{tab:model-performance} compares the performance of three different deep learning architectures—VGG16, AlexNet, and CNN—on the Indian Sign Language (ISL) dataset. The CNN model achieved the highest accuracy of 99.95\%, significantly outperforming both VGG16 (95.56\%) and AlexNet (93.9\%). This comparison highlights the superior effectiveness of the CNN in accurately recognizing ISL gestures, demonstrating its ability to learn complex patterns and features from the dataset.
\section*{Conclusion}
This research has developed a CNN-based system for detection of  Indian Sign Language (ISL) gestures, contributing to the field of accessible communication technology. The model demonstrated high accuracy and robust performance across various ISL classes, effectively handling the complexities of hand gestures. Key image preprocessing techniques, including thresholding, edge detection, and skeleton morphology, played a significant role in enhancing the model's feature extraction capabilities. Furthermore, data augmentation methods helped mitigate overfitting, making the model resilient to variations in gesture orientation and environmental lighting. Comparative analysis with traditional classifiers confirmed the superiority of CNNs for visual classification tasks, showcasing the value of deep learning in addressing sign language recognition challenges. This research establishes a strong foundation for implementing real-time ISL recognition systems, offering a promising tool for enhancing communication accessibility for the hearing and speech impaired.

\bibliographystyle{unsrt}  
\bibliography{references}  

\end{document}